# Synthetic data generation for Indic handwritten text recognition

[a]Partha Pratim Roy*, [b]Akash Mohta, [c]Bidyut B. Chaudhuri

[a]Dept. of CSE, Indian Institute of Technology Roorkee, India
[b]Dept. of ECE, Institute of Engineering & Management, Kolkata, India
[c]Computer Vision & Pattern Recognition Unit, Indian Statistical Institute, Kolkata, India
[a]email: proy.fcs@iitr.ac.in, TEL: +91-1332-284816

## Abstract

This paper presents a novel approach to generate synthetic dataset for handwritten word recognition systems. It is difficult to recognize handwritten scripts for which sufficient training data is not readily available or it may be expensive to collect such data. Hence, it becomes hard to train recognition systems owing to lack of proper dataset. To overcome such problems, synthetic data could be used to create or expand the existing training dataset to improve recognition performance. Any available digital data from online newspaper and such sources can be used to generate synthetic data. In this paper, we propose to add distortion/deformation to digital data in such a way that the underlying pattern is preserved, so that the image so produced bears a close similarity to actual handwritten samples. The images thus produced can be used independently to train the system or be combined with natural handwritten data to augment the original dataset and improve the recognition system. We experimented using synthetic data to improve the recognition accuracy of isolated characters and words. The framework is tested on 2 Indic scripts - Devanagari (Hindi) and Bengali (Bangla), for numeral, character and word recognition. We have obtained encouraging results from the experiment. Finally, the experiment with Latin text verifies the utility of the approach.

# 1. Introduction

Today, large amount of information is stored in the form of physical data, that include books, handwritten manuscripts, forms etc. Documents present in physical forms need to be converted to digitized format for easy retrieval and usage. This digitization task is not an easy one and requires significant amount of labor to convert the documents to digital format. Handwriting recognizers are being used to digitize such information. Until now, only a limited amount of research has been done on handwritten Indic script text recognition. A lot of Indic scripts remain, for which no formal datasets have been created. Due to this dearth of datasets, it becomes extremely hard to train systems for recognizing such scripts.

Experiments in the field of pattern matching show that the recognition performance does not depend only on the features and classification algorithms, but also on quality and size of the training data. Researchers interested in creating recognition systems need to invest a lot of effort in creating datasets of target script for training purpose. However training data is not readily available for every language. This creates a problem for such systems, as the only alternative that remains is to create the data manually. This is a very tedious and time consuming process, as texts and labels need to be created simultaneously. The solution to this problem is to create synthetic data automatically. However, creation of synthetic data requires extensive user interference. Present systems of synthetic data creation are customized to the script in question. Each time synthetic data for a new script needs to be generated, a new system needs to be created that requires management on minute scales. The system is nearly rendered useless after the creation of datasets. Moreover, most of such systems require initial natural handwritten data to work upon.

Although, a number of work [23, 35] have been developed in this area to expand the natural dataset but they still require natural dataset to work upon. Style-preserving handwriting synthesis [24] remains as an option, but owing to the need of stroke analysis and sample collecting interface, human interference and input is required. Existing research work for handwriting synthesis can be divided broadly into two categories. The first category is based on the reconstruction process [27, 34] in which the handwriting stroke-trajectory is modeled by velocity or force functions. This requires human input and an online system that records the velocity and force variations. This method is usually not preferred for non-cursive styled languages. The second category uses glyphs [22, 37, 38] where the systems record the handwritten elemental symbols directly and reuse them during synthesis. Methods in this category thus require user involvement in the sample collection process.

The process of creating synthetic dataset without any human input still remains an unexplored area. This kind of dataset can significantly reduce the time and resources spent on creating a natural dataset. Fig.1 illustrates the block diagram of text recognition system using synthetic/natural dataset. Feature extraction and training is performed from manual labeling of dataset images. Once the character models are trained using classifiers, they can be used for testing unknown characters/word images. Training of character models can also be performed using synthetic dataset. We can also use synthetic deformation of naturally available handwriting samples and improve the classifier for recognition system.

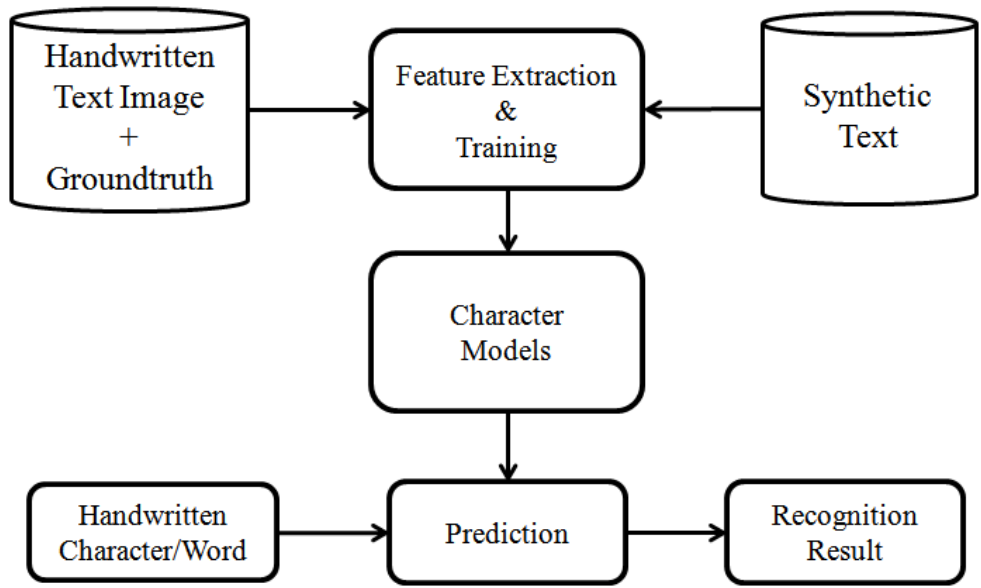

**Fig.1. Block diagram of the text recognition system using natural and synthetic dataset.**

## 1.1. Related Work

There are two primary approaches to handwriting synthesis, namely top-down [12, 16] and bottom-up [38]. Top down approaches use neuromuscular action models to replicate natural writing motion whereas bottom-up approach models written samples. The top-down approach requires online movement of data which can be recorded by tablets; hence bottom-up approaches can be deemed more practical for offline recognition.

Several methods can be used for synthesis of synthetic data. Generation techniques produce new images using input samples. New images produced are at the same level as that of input sample i.e. new character sample from input character sample. Concatenation techniques use basic glyphs to produce samples consisting of more than one glyph. In [33], use of character template was proposed to reduce human input. The character templates that represent the ideal shapes of the different letters, were built for upper and lower case of English alphabet, for the ten digits and special symbols. Each prototype was put together manually from a series of Bezier arcs, in a predefined manner.

Perturbation based techniques use geometric deformation on natural writing to augment training sets of recognition systems [7, 8, 13, 35]. The advantage of this perturbation model is that a limited set of natural writing could be extended to create a larger dataset with varied features for training. This model generally takes text lines as input from different writers. Variation in human writing occurs due to many sources such as letter shape, writing instrument and others. The perturbation model uses parameters with a range of values from which a random value is picked every time for distorting the line. But, before applying such distortions, the image needs to be pre-processed to be geometrically true to the distortion values. The distortions need to be done in a non-linear fashion so that they are not reversed by standard linear pre-processing operations performed by general handwriting recognition systems. However, perturbation based models [7] still require human inputs to work upon, which demands efforts towards data collection.

Lina and Wanb [24] proposed an approach to synthesize handwriting according to user's writing style. In this system, handwritten features such as character glyph, size, slant, ligature style, character spacing and cursiveness are extracted to understand user's writing style. This approach requires the user to input each character, some special pairs of characters and multi letter words. Next, the system synthesizes handwriting hierarchically using writing style. Each glyph is modified and aligned on the baseline with feasible distance. Adjacent letters are connected by polynomial interpolation. Words are then rendered one by one to form a line. This method though keeps the dataset close to natural handwriting, requires a great amount of sample from a single user. This defeats the purpose of creating synthetic dataset with minimized human effort. Kinematic Theory of rapid human movements [17, 18] has also been explored in order to generate totally synthetic specimens. Researchers [16, 17] have used this technique for the generation of synthetic on-line signatures.

There exist a number of work [11, 14, 32] in Arabic script for synthetic text generation. In [32], the authors proposed a system that automatically generates prototypes for each word in a lexicon using multiple appearances of each letter. Large sets of different shapes are created for each letter in each position. However, not much work is available for synthetic data generation in Indic scripts. The OCR for recognizing printed Devanagari and Bangla scripts has been discussed in many pieces of research work [26]. Although a number of work has been investigated for handwritten numeral and character recognition in Indian script [4], only few papers are available towards handwritten word recognition in Indian script [5]. One of the main reasons is not availability of proper dataset for training recognition system. Offline recognition of handwritten texts of these scripts needs lot of research.

In this paper we present a method to generate synthetic handwritten data without usage of natural writing to train recognition systems. The idea of the approach proposed in this paper is to use the available fonts and generate the words. Then two different deformation techniques, namely vectorization-based distortion [10] and image-based deformation [20] are performed on the generated word images. Distortion using vectorization was performed on skeleton information of the text image. Other deformations are performed by curving the word, modifying the size of the characters, etc. To the best of our knowledge, the proposed approach has not been used earlier for synthetic data generation. In our approach, the system fonts are distorted to generate a dataset with a large variety in data. Though the proposed distortion models do not produce new printed/handwritten text, but can modify the existing text to a certain degree to help recognizer for improving performance. The main advantage being that it neither involves human input nor handwritten samples. This present work mainly deals with Indic scripts, which can be extended to other scripts, provided the availability of proper fonts. The datasets for two scripts have been synthetically generated, namely Devanagari and Bengali. The system has been tested on handwritten characters after subsequent training using Support Vector Machines (SVM) [36]. For word recognition, the classification is performed using Hidden Markov Model (HMM) [18].

The rest of the paper is organized as follows. In Section II we talk about properties and challenges in Indic script recognition. In section III, we present the method for generation of synthetic handwritten data and the steps involved in generation of the curved words. Section IV demonstrates the results obtained after testing with synthetically generated images for recognition of handwritten images. Section V presents the conclusion and scope of future work in this field.

## 2. Properties and Challenges in Bengali and Devanagari Scripts

Devanagari (Hindi) is the most widely used Indic script [26] and is used in writing many scripts, such as Sanskrit, Devanagari, Marathi, and many others. In Devanagari script, a total of 49 basic characters exist, out of these 11 are vowels and 38 are consonants. Bengali (Bangla) is the second most popular language in India and the fifth most popular language in the world [26]. About 200 million people in the eastern part of India speak this language. The alphabet of the modern Bengali script consists of 11 vowels and 39 consonants. Some text examples of Bengali and Devanagari scripts are shown in Fig.2. There are no distinct upper and lower case letter forms. It is seen that most of the characters in Bengali and Devanagari scripts have a horizontal line (*Matra/Shirorekha*) at the upper part and a baseline. When two or more characters are written side by side they form a word. All Indic scripts run from left to right, although some glyphs, while combine, appear to the left of their base character. In both Bengali and Devanagari scripts, a vowel following a consonant takes a modified shape and is placed at the left, right, both left and right, or bottom of the consonant. These modified shapes are called *modified characters*. These modifiers add more difficulties in the character segmentation procedure because of their topological position. A consonant or a vowel following a consonant sometimes takes a compound orthographic shape, and these are called as *compound character*. For details about Bengali and Devanagari scripts, we refer [26].

A Bengali or Devanagari text can be partitioned into three zones. The *upper-zone* denotes the portion above the *Matra*, the *middle zone* covers the portion between *Matra* and base-line, the *lower-zone* is the portion below the base-line. Like other languages, characters in these also suffer from distortions which depend on the writing style of the individual. Hence, character segmentation from a word may often fail due to the presence of touching characters, noise etc. Due to improper segmentation process, characters may generate disjoint character components which may create problems in the recognition tasks. Proper classification of these components using segmentation is not easy to process. Also, non-uniform skew and slant in word images make the recognition task difficult. As mentioned earlier, "Matra" stays in a horizontal line dividing the upper and mid-section of the text, which often do not follow straight line in handwritten word images.

**(a)**

**(b)**
**Fig.2. Sample text of (a) Devanagari and (b) Bengali script taken from [28]**

Recognition of Bengali and Devanagari scripts is not similar to Latin due to the variation of character-modifiers present in 3 zones, upper, middle and lower as depicted in Fig.3. When the consonant character, "ক" (appear only in middle zone) gets combined with a vowel, the vowel forms a modifier which can appear in either the middle zone (like "কা"), or the middle and upper zones (like "কি"), according to the nature of the vowel. Hence, the conjunction of consonants and vowels produce a large number of possible character combinations. To overcome the problem of huge number of character recognition, it was shown that zone-wise recognition method improves the word recognition performance than conventional full word recognition system in Indic scripts [31].

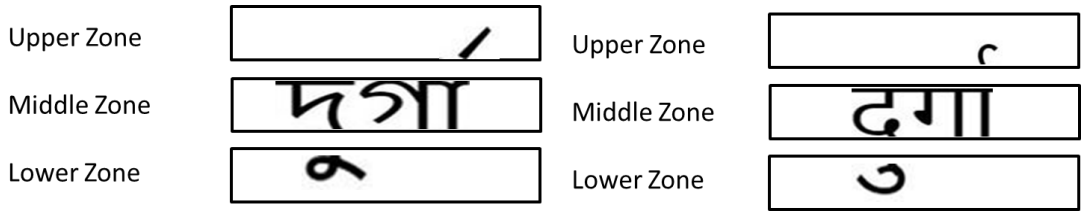

**Fig.3. Division of word image into upper, middle and lower zones. (a) Bengali word (b) Devanagari word.**

## 3. Synthetic Text Generation

The proposed method can generate dataset of images in any script, thus proving it to be a powerful tool for generating large amount of training data for scripts whose handwritten data is not easily available. The images of words are generated using multiple fonts from digital data. The images so produced can be deformed such that it makes the characters different in size or shape like most of the modern day writing. For this purpose, the images are vectorized and next vectors are distorted to produce handwritten-like images. Also, the images can be deformed without vectorization by applying size/skew deformation on different parts of the image. Finally, both types of distortion models can be combined to augment dataset. Fig.4 shows the flowchart of the synthetic data generation system from text collected from electronic media (e-media).

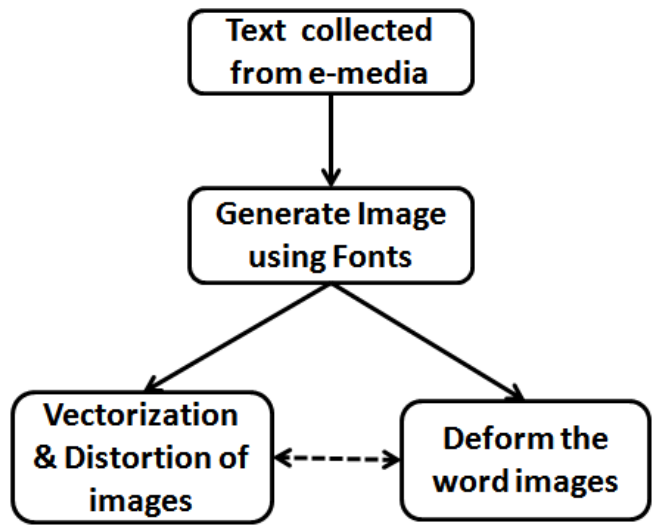

**Fig.4. Block diagram of synthetic word generation approach**

## 3.1. Generation of Images with Font Variation

To obtain the synthetic images of the target script, the images with the required labels are generated using varying styles of characters. These styles are obtained from different fonts of corresponding script. A number of open source libraries are available to generate text characters of different fonts. We have used ImageMagick toolbox [1] due to its handy usage. Some examples of numerals, characters and words of English, Devanagari and Bengali scripts generated by this toolbox are shown in Fig. 5.

| English | 1 2 3 4 | a b c d | India |
| Devanagari | १ २ ३ ४ | क ख ग घ | भारत |
| Bengali | ১ ২ ৩ ৪ | ক খ গ ঘ | ভারত |

         (a)      (b)      (c)

**Fig.5: Images generated using ImageMagick. (a) Numerals, (b) Characters, and (c) Words**

## 3.2. Vectorization and Distortion

The images created are vectorized using algorithm due to Rosin and West [30]. For this purpose, the image is first skeletonized [33]. Next, the skeleton segments are vectorized using polygonal approximation [30]. The principle of this vectorization approach is to recursively split the curve into small segments, until the maximum deviation is zero or the deviation is less than three pixels. Next, the branching of possible segments is traced such that the segments maximizing a measure of significance are stored. The significance is measured by a ratio of the maximum deviation and the length of the segment.

The vectorization of images is performed using image processing toolbox QgarApps [28]. To obtain the vectorized image, the initial and final points of the vectors obtained are connected using the Bresenham line generation algorithm (See Fig. 6). Distortion in vectors is explained in following sub-sections.

| Language | Fonts 1 | | Fonts 2 | | Fonts 3 | |
|---|---|---|---|---|---|---|
| | Actual | Vectorised | Actual | Vectorised | Actual | Vectorised |
| Bengali Numerals | ৮ | ৮ | ৮ | ৮ | ৮ | ৮ |
| Devanagari Character | भारत | भारत | मारत | मारत | भारत | भारत |
| Devanagari Word | क | क | क | क | क | क |

**Fig.6: Vectorization of Images**

The vector points obtained after vectorization of images are used to bring the required distortion in the images. The coordinates of the points are shifted to produce a variety of images that consort to handwritten images. In our framework we have used two different ways to do so; random shift of segments and controlled distortion using Gaussian distribution.

**a) Random Shift of Segments:** Using this approach, the $x$ and $y$ coordinates of the vector points are shifted to different locations to modify the original image. The extent of shift in the images depends on the resolution of the image [19]. A random number generator was used to create the distortions. The shifts in this approach are produced according to the following formula:

$$x' = x + 0.05 \times w \times r \quad (1)$$
$$y' = y + 0.05 \times h \times r \quad (2)$$

where, $w, h$ are width and height of image in pixels, and $r$ is a random number ranging from -1 to +1. The shift of the point from original position (x, y) to a new position (x', y') is measured by Euclidean distance using Eq. (3).

$$d = \sqrt{(x-x')^2 + (y-y')^2} \quad (3)$$

The variation in images with maximum value of $d$ in pixels is shown in Fig. 7.

| Distortion / Language | 0 pixels | 6 pixels | 8 pixels | 10 pixels | 12 pixels | 14 pixels |
|---|---|---|---|---|---|---|
| Bengali Numerals | ৮ | ৮ | ৮ | ৮ | ৮ | ৮ |
| Devanagari Character | क | क | क | क | क | क |
| Devanagari Word | भारत | भारत | भारत | भारत | भारत | भारत |

**Fig.7: Images generated using distortion of vectorized images.**

**b) Controlled Distortion using Gaussian Distribution:** A controlled distortion in segment shifting was performed using Gaussian distribution [20]. The probability distribution of Gaussian distribution is

$$p(x) = \frac{1}{\sqrt{2\pi\sigma^2}} e^{-\frac{(x-\mu)^2}{2\sigma^2}}, \quad (4)$$

where, $\mu$ and $\sigma$ are the mean and standard deviation respectively. The function has its peak at the mean, and "spread" increases with the standard deviation. This implies that distortion based on Gaussian function is more likely to return samples lying close to the mean (initial point), rather than those far away. To implement distortion using Gaussian distribution, we set the mean value as the initial vertex derived after vectorization of the image. The value of $\sigma$ is controlled by the user and marks the maximum deviation that can be obtained on any vertex point. By using Gaussian distributions the probability of large excitation of image is reduced. This helps in reducing the number of images having very high distortional levels while training. We experimented with varying values of $\sigma$, the best results were obtained when $\sigma$ was set as $0.02 \times Min(h,w)$. Where $Min(h,w)$ computes the minimum of height and width the text image.

The vectorization process segments the images into polylines at the junction points. Therefore shifting the junction points randomly relocates the point that causes discontinuities in images sometimes.

Hence, a correction was applied to preserve the structure of images. To do so, the endpoints of distorted vectors were connected using the Bresenham algorithm (See Fig. 8).

| Script | Random Shifting | Gaussian Distribution |
|---|---|---|
| Devanagari | भारत क | भारत क |
| Bengali | তাতে প | তাতে প |

**Fig.8: Words and characters of Devanagari and Bengali scripts are generated by random shift of segments and controlled distortion using Gaussian distribution.**

**3.3. Image Deformation without Vectorization**

As discussed earlier, in most of the handwritten samples, characters are not uniform in size or shape to each other. Most of these samples appear either skewed or curved in a specific manner. In this section, we developed three types of deformations, curved, sinusoidal and elliptical to generate synthetic images. These are detailed as follows.

**a) Curved Distortion:** The word images generated using ImageMagick [1] is divided into different segments column wise [20]. These various segments so obtained are moved either up and down in order to generate a curved text line in nature. In our experiment we have generated curved text lines of two kinds, rainbow and inverted rainbow shapes. The images obtained after curved distortion are shown in Fig.9. To remove discontinuity between two segments, shear transformation is applied for smoothening.

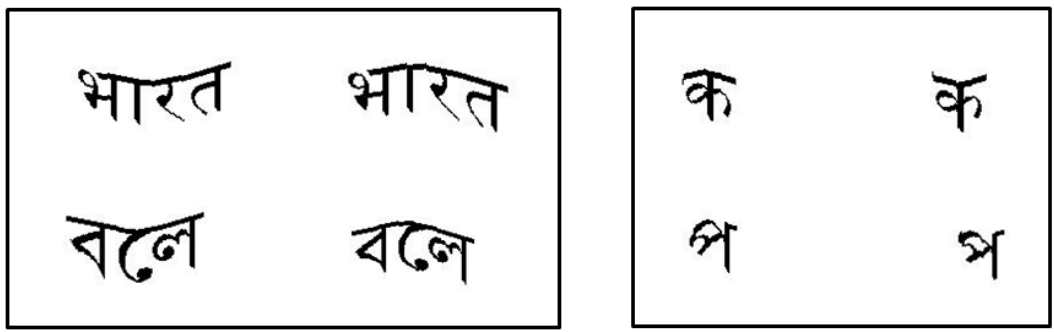

**Fig.9: Images generated using curved distortions**

**b) Sinusoidal Distortion:** The next type of distortion is in the form of a sinusoidal curve [20]. The images are divided into various segments column wise similar to curved distortion. Next, the segments are moved in up and down directions, thus producing the shape of a sinusoid as shown in Fig.10. Next, the staircase effects between segments are smoothened by applying shear transformation according to their orientation.

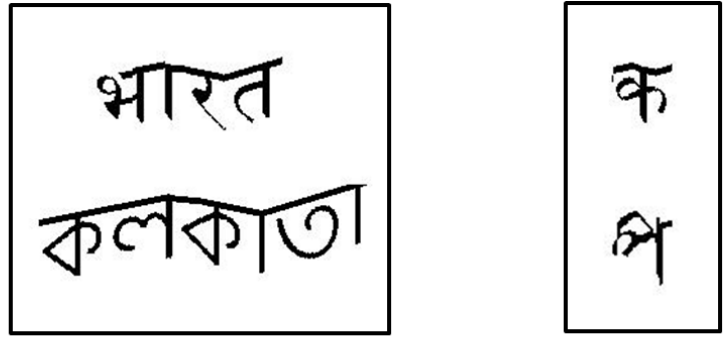

**Fig.10: Sinusoidal Distortion in Bengali and Devanagari words (left) and characters (right).**

**c) Elliptical Distortion:** Handwritten samples may contain characters which are different in size to the rest in the text [15]. To resemble such cases, we have used an elliptical type of distortion where the middle portion of the text is larger in size than the rest of the text and as a result the total word appears in the form of an ellipse [20]. To achieve this distortion type, we use equation of ellipse given in Eq. (5).

$$\frac{(x-x_c)^2}{a^2} + \frac{(y-y_c)^2}{b^2} = 1 \qquad (5)$$

where $(x_c, y_c)$ is the center of the ellipse. a and b are the lengths of the semi major axis and semi minor axis respectively. We then fit the word to that equation and thus obtain the middle portion of the word larger in size than the rest. The words obtained after elliptical distortion are shown in Fig.11.

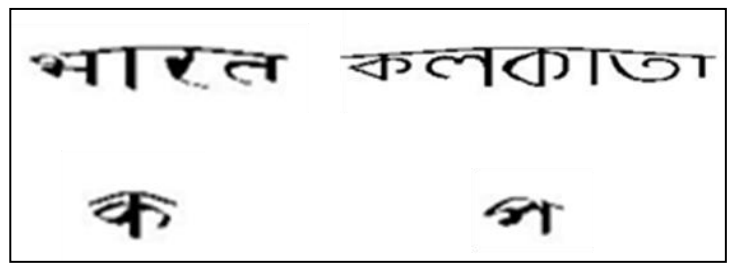

**Fig.11: Shows elliptical distortions in Bengali and Devanagari Words**

### 3.4. Combination of Image Deformation and Vectorization

The words generated after individual distortion models bear little resemblance to handwritten images. In order to make a close appearance with actual handwritten samples we apply the vectorization distortion (as discussed in Section 3.2) on these deformed images made by shape deformation. Fig.12 shows the distorted images after applying deformation and vectorization distortion.

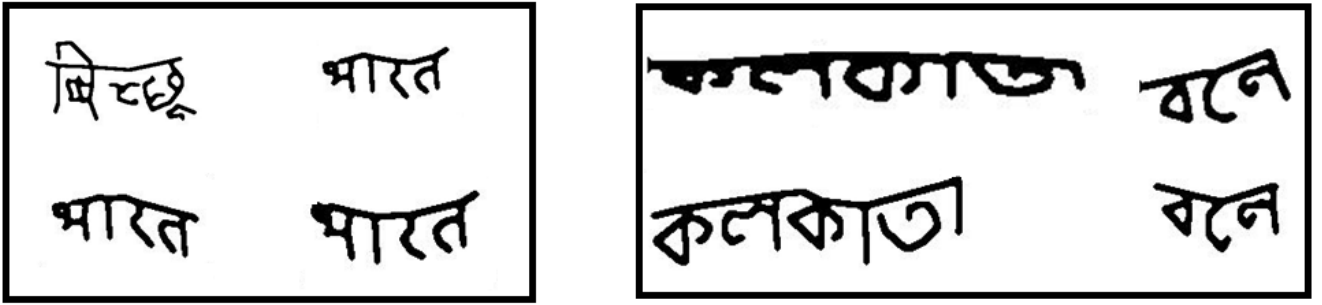

**Fig.12: Shows some of the distorted words after undergoing shifting and correction**

## 4. Text Recognition Framework

### 4.1. Isolated Character/Numeral Recognition

N. Das et al. [9] provided a Principal Component Analysis (PCA) based feature for isolated numeral recognition. We have considered similar approach to test the recognition performance by synthetic isolated data generation. The approach is discussed below.

#### 4.1.1. Feature Extraction

PCA [6, 21] is a popular dimension reduction model used in various character recognition approaches. It is an eigenvector based technique that utilizes variance of data. To extract PCA-based features, each grey level character/numeral image is first converted to minimum bounding box and next they are normalized to the size of $M \times M$ ($=N$) pixels. Next, $P$ principal components are chosen corresponding to top $P$ (P < N) Eigen vectors from $v_j$ by solving the following Eq (6).

$$Av_j = \lambda_j v_j, \text{ j=1, 2, ...,N} \quad (6)$$

Where $A$ is the covariance matrix and $\lambda_j$s are the Eigen values corresponding to Eigen vectors. For recognition, the grey images are first resized to 150x150. Next, top $P$ principal components are extracted from these images and considered for classification.

### 4.1.2. Classification using SVM

Support Vector Machine (SVM) classifier [8] has been used to classify isolated characters/numerals. SVM classifier has been chosen as it has successfully been applied in a wide variety of classification problems [8]. Given a training database of *M* data: $\{x_m / m=1,..,M\}$, the linear SVM classifier is defined in Eq. (7) as:

$$f(x) = \sum_j \alpha_j x_j + b \qquad (7)$$

Where, $x_j$ is the set of support vectors and the parameters $\alpha_j$ and $b$ have been determined by solving a quadratic problem. A linear kernel can be used to classify data which have fewer variations. But changing the kernel function to Radial Basis Function (*RBF*) was a better choice in our recognition system. The Gaussian kernel is of the form given by Eq. (8) where $\sigma$ is the standard deviation.

$$K(x, x') = \exp\left(-\frac{\|x - x'\|^2}{2\sigma^2}\right) \qquad (8)$$

### 4.2. Word Recognition

Recently, it was shown that zone-wise recognition method in Indic scripts [31] improves the word recognition performance than conventional full word recognition system. It is due to the fact that zone segmentation significantly reduces the combination of characters with the modifiers which provides effective increase in recognition performance than considering full characters for recognition. We discussed some advantages of zone segmentation based Indic text recognition in Section 1. Here, we follow the zone segmentation approach for Indic text recognition and show that synthetic data can improve the recognition performance further. Feature extraction and recognition process for word recognition are detailed here.

### 4.2.1. Feature Extraction

Pyramid Histogram of Gradient (PHOG) [25] is the spatial shape descriptor which provides the feature of the image by spatial layout and local shape, comprising of gradient orientation at each pyramid resolution level. To extract the feature from each sliding window, we have divided it into cells at several pyramid level. The grid has $4^N$ individual cells at *Nth* resolution level (i.e. *N*=0, 1, 2, ...). Histogram of gradient orientation of each pixel is calculated from these individual cells and is quantized into *L* bins. Each bin indicates a particular octant in the angular radian space. The concatenation of all feature vectors at each pyramid resolution level provides the final PHOG descriptor. *L*-vector at level zero represents the *L* bins of the histogram at that level. At any individual level, it has $Lx4^N$ dimensional feature vector where *N* is the pyramid resolution level (i.e. *N*=0, 1, 2….). Thus, the final PHOG descriptor consists of $L \times \sum_{N=0}^{N=K} 4^N$ dimensional feature vector, where *K* is the limiting pyramid level. In our implementation, we have limited the level (*N*) to 2 and we considered 8 bins (360º/45º) of angular information. So we obtained (1×8) + (4×8) + (16×8) = (8+32+128) = 168 dimensional feature vector for individual sliding window position.

### 4.2.2. HMM based Recognition

We extracted PHOG feature descriptors using sliding window and applied HMM for word recognition [3, 31]. One of the important features of HMM is the capability to model sequential dependencies. In our recognition system, words are modelled by the concatenation of its middle zone component models. Bakis topology with continuous observation densities has been used for word recognition. The observation probability density for each state is a mixture of Gaussian distributions. This mixture is obtained by incrementing (with a step of power of 2) the number of Gaussian distributions in each state until a convenient HMM topology is found satisfactory. The HMM models are initialized with one Gaussian distribution per state and are trained using Baum-Welch algorithm.

An HMM can be defined by initial state probabilities $\pi$, state transition matrix $A = [a_{ij}]$, $i, j=1,2,...,N$, where $a_{ij}$ denotes the transition probability from state *i* to state *j* and output probability $b_j(O_K)$ modelled with continuous output probability density function. The density function is written as $b_j(x)$, where *x* represents *k* dimensional feature vector. A separate Gaussian mixture model (GMM) is defined for each state of model. Formally, the output probability density of state *j* is defined as:

$$b_j(x) = \sum_{k=1}^{M_j} c_{jk} \mathcal{N}(x, \mu_{jk}, \Sigma_{jk}) \quad (9)$$

where, $M_j$ is the number of Gaussians assigned to *j*. and $\mathcal{N}(x, \mu, \Sigma)$ denotes a Gaussian with mean $\mu$ and covariance matrix $\Sigma$ and $c_{jk}$ is the weight coefficient of the Gaussian component *k* of state *j*. For a model $\lambda$, if *O* is an observation sequence $O = (O_1, O_2,..., O_T)$ which is assumed to have been generated by a state sequence $Q = (Q_1, Q_2, ..., Q_T)$, of length $T$, we calculate the observations probability or likelihood as follows:

$$P(O, Q | \lambda) = \sum_Q \pi_{q_1} b_{q_1}(O_1) \prod_T a_{q_T-1\ q_T} b_{q_T}(O_T) \quad (10)$$

Where $\pi_{q_1}$ is initial probability of state 1.

The recognition is performed using the Viterbi algorithm. Viterbi algorithm produces a sequence of states and corresponding score to generate hypothesis result of middle zone. For a given middle-zone of a word image (X), its score is calculated based on a lexicon (W) of the middle zone characters and it is the posterior P(W|X). Using logarithm in Bayes' rule we get

$$\log p(W|X) = \log p(X|W) + \log p(W) - \log p(X) \quad (11)$$

From these scores N-best hypotheses are chosen. Now among these N-best choices, the best hypothesis is chosen combining upper and lower zone information. It is discussed in following subsection.

**4.2.3. Indic Text Recognition using Zone Segmentation**

As mentioned in [31], we performed a 'Matra' detection based upper zone separation technique and projection profile based lower zone separation method for word images. We apply HMM [18] for recognizing the touching components in middle zone. The HMM is used for its capability to model sequential dependencies. During the training phase, the transcriptions of the middle zone of the word images are used to train the character models. The isolated components which are present in upper and lower zones are segmented using connected component (CC) analysis and labelled as text characters. For classification, the images are resized into 150x150 and then PHOG features of 168 dimensions are extracted from upper and lower zone modifiers. PHOG feature is considered as it provides better result in the experiment. Support Vector Machine (SVM) classifier [8] has been used to classify these components. The *Radial Basis Function (RBF)* kernel function was used in our case because of its better performance. The upper and lower modifiers have been separated by the alignment information as discussed in the following subsection.

For estimating the boundaries of the characters in Indic word, Viterbi forced alignment *(FA)* has been used. With embedded training, the optimal alignments of a set of HMMs are found. The results of alignment are mapped to modifiers in upper and lower zones. In the middle zone, after alignment of the character segments, multiple hypotheses are obtained (see middle zone alignment in Fig.13). As discussed earlier, we generate *N*-best Viterbi list composed of *N* hypotheses. *N*-best lists are generated to obtain a set of likely middle-zone word hypotheses. These associate different labeling and segmentation pairs. Every pair of segmentation and label in the list is given a confidence measure by HMM log-likelihood. Among all, the best word hypothesis can be chosen based on combining recognition results of the modifiers from upper and lower zones [31]. Finally, the zone-wise recognition results are combined to form the whole word (see Fig.13).

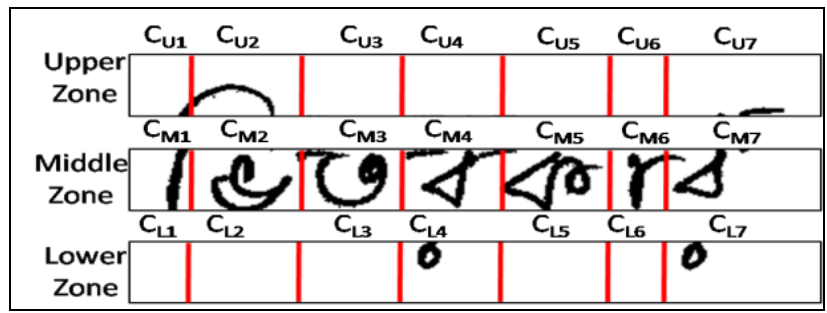

**Fig.13.Middle zone alignment result applied to upper and lower zones for modifier separation purpose. All extracted regions are fed in SVM for classification.**

## 5. Experimental Results & Discussions

We have tested the synthetic text generation scheme on numerals, characters and words. Recognition of isolated characters and numbers is performed using SVM and the word recognition is done using HMM. The experimental results are reported in following subsections.

**5.1. Isolated Character and Numeral Recognition**

We tested our system for numeral recognition as well as character recognition in 2 different Indic scripts: Bengali and Devanagari. Dataset details and experimental evaluation are discussed in following.

**5.1.1. Dataset Description**

For training purpose, we generated synthetic images of Devanagari and Bengali scripts. We used 25 fonts for each script. Some of these font names are mentioned in Table I. The font size for numerals and characters are chosen to be 150. From each generated image we produced 200 distorted images using distortions with and without vectorization. For the testing process standard images from CMATER (Center for Microprocessor Applications for Training Education and Research) [3] database have been used. It provides a comprehensive Indic script character database. Table II shows the number of images used for training and testing for numerals and character recognition. Fig. 14 shows the testing data obtained from CMATER database.

**Table I: Some fonts used for synthetic data generation**

| Devanagari Fonts | Bengali Fonts |
|---|---|
| DevLys-310 | Ekushay Bengali |
| Kruti-Dev-010 | Ekushay Kolom |
| Kruti-Dev-012 | Ekushay Sumon |
| Kruti-Dev-013 | Ekushay Lal Sabuj |
| DevLys-150 | Kalpurush |
| Dina-15 | Siyam Rupali |
| Rukmani-Regular | Mitra Mono |
| Safaltaa | Mukti |
| Mangal | Bensen |
| Kruti-Dev-678 | Aponalohit |

**Table II: Data details used in performing experiment for numeral and characters.**

| Script | Type | Experiment | Training | Testing |
|---|---|---|---|---|
| Devanagari | Numeral | Synthetic data | 30000 | 3000 |
| | | Synthetic + handwritten | 34000 | 3000 |
| | | Synthetic + handwritten (deformed) | 54000 | 3000 |
| | Character | Synthetic data | 38000 | 5000 |
| | | Synthetic + handwritten | 42000 | 5000 |
| | | Synthetic + handwritten (deformed) | 62000 | 5000 |
| Bengali | Numeral | Synthetic data | 30000 | 4000 |
| | | Synthetic + handwritten | 33500 | 4000 |
| | | Synthetic + handwritten (deformed) | 52000 | 4000 |
| | Character | Synthetic data | 38000 | 6000 |
| | | Synthetic + handwritten | 41000 | 6000 |
| | | Synthetic + handwritten (deformed) | 59000 | 6000 |

| English | 0 | 1 | 2 | 3 | 4 | 5 | 6 | 7 | 8 | 9 |
|---|---|---|---|---|---|---|---|---|---|---|
| Devanagari | ० | ९ | २ | ३ | ४ | ५ | ६ | ७ | ८ | ९ |
| Bengali | ০ | ১ | ২ | ৩ | ৪ | ৫ | ৬ | ৭ | ৮ | ৯ |

**(a)**

| Devanagari | अ | आ | इ | ई | क | ख | ग |
|---|---|---|---|---|---|---|---|
| BENGALI | অ | আ | ই | ঈ | ক | ধ | গ |

**(b)**

**Fig.14: Examples of handwritten test data (a) Numerals (b) Handwritten Characters**

### 5.1.2. Font wise recognition results

We have tested with some of these fonts individually for character and numeral recognition. Fig.15 provides recognition performance of each font in Bengali and Devanagari datasets. Synthetic data was generated using distortion in each font. The disparities in accuracy arise due to the difference in the pattern/style among fonts. Hence, one can conclude that proper usage of fonts can provide higher result in recognition. By combining data of all fonts we achieved recognition accuracy of 77.21% and 75.32% in Devanagari and Bengali character sets, respectively. The individual font wise accuracy is less than the overall accuracy because the recognition is dependent on the number of samples provided for training.

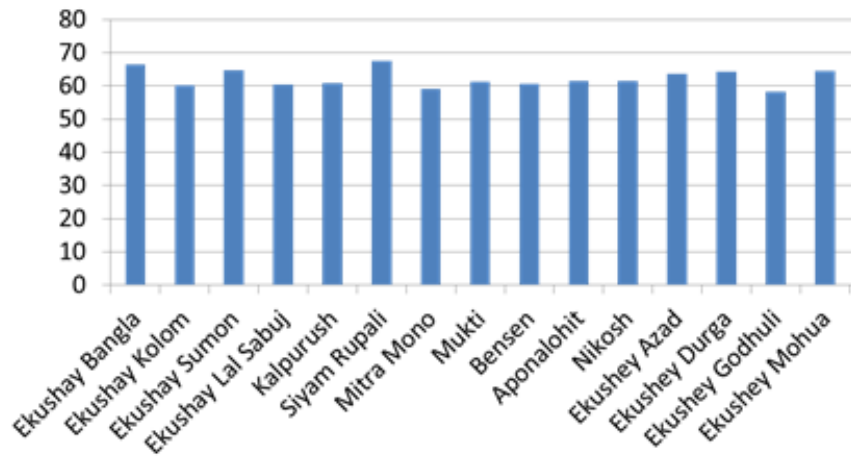

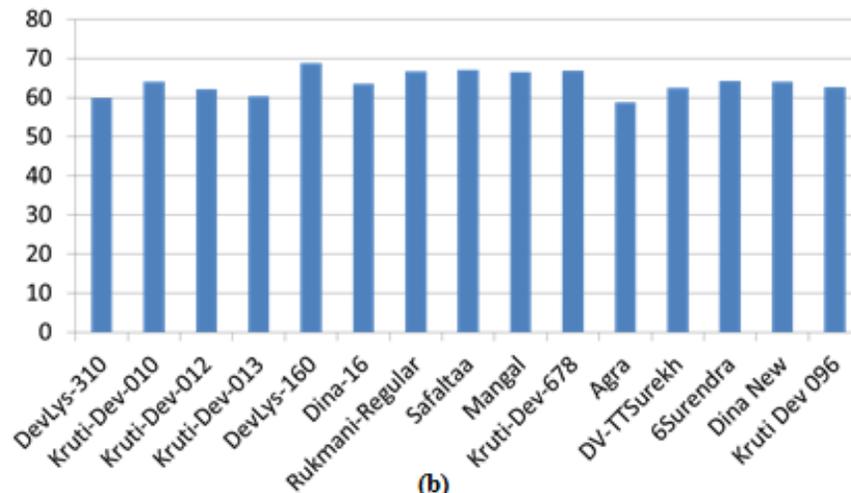

**Fig.15: Performance evaluation of handwritten characters in (a) Bengali (b) Devanagari scripts.**

**5.1.3. Recognition Performance with Distortion Models**

As discussed earlier, various distortions were performed on the generated synthetic data. Individually synthetic text generated from each distortion model i.e. curved deformation, elliptical distortion, etc. were taken and tested for recognition in characters to give an insight about the effect of the various distortion models. Finally, all different types of fonts and deformation models etc. were combined and overall two types of recognition was performed. Firstly, only synthetic data was considered in training and the result was tested on actual handwritten samples. Secondly, synthetic data was combined with handwritten samples for training and then tested for recognition. Fig.16 shows results with different distortions in isolated characters and numerals in Devanagari and Bengali scripts. We noted that distortion due to curve, elliptical images gave us better results. Vectorization distortion with random shift provided better results than Gaussian distortions.

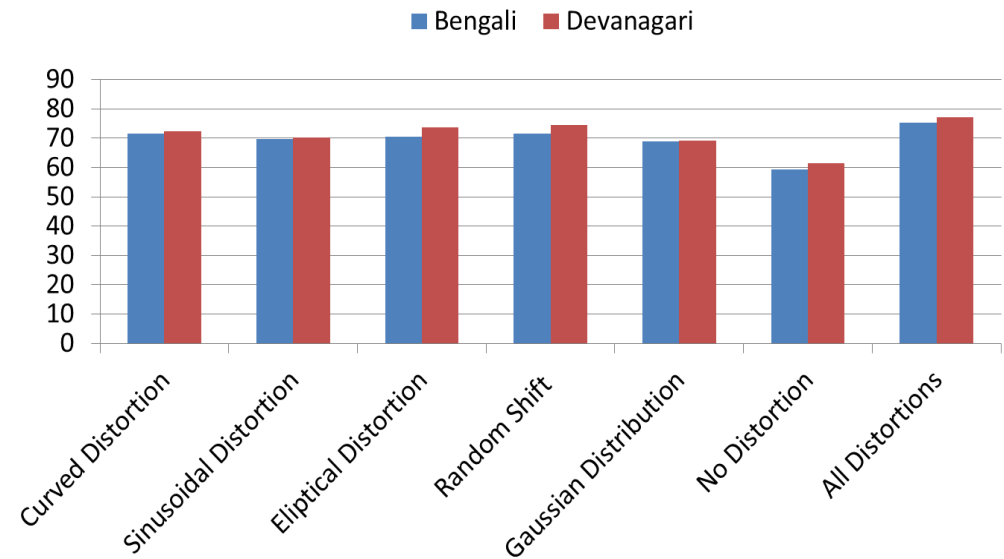

(a)

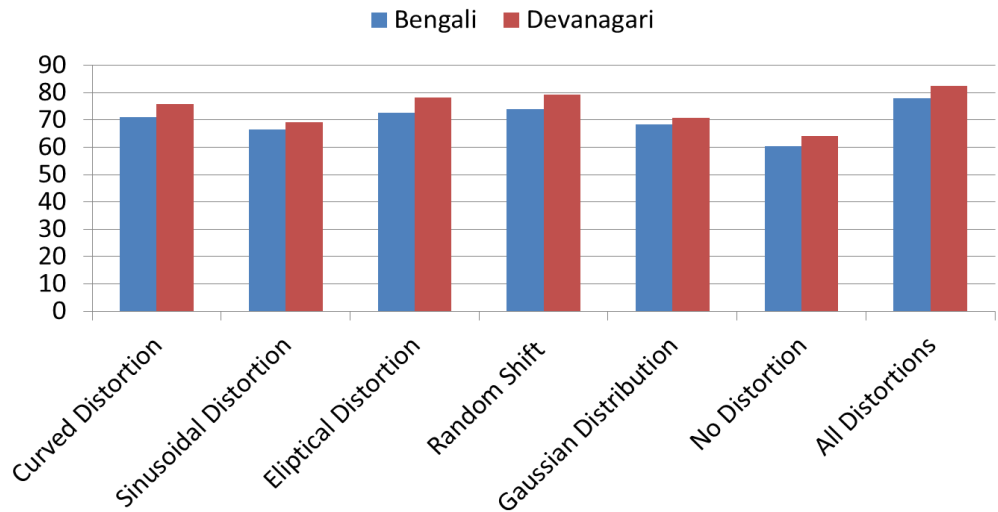

(b)

**Fig 16: Recognition of characters with different distortions. (a) Character (b) Numeral.**

The accuracy of the system heavily depends on the variation of writing styles we cover in our training dataset i.e. total number of different fonts we use while training. We found that the accuracy of the system improves as we increase the number of fonts used for training the system (See Fig. 17).

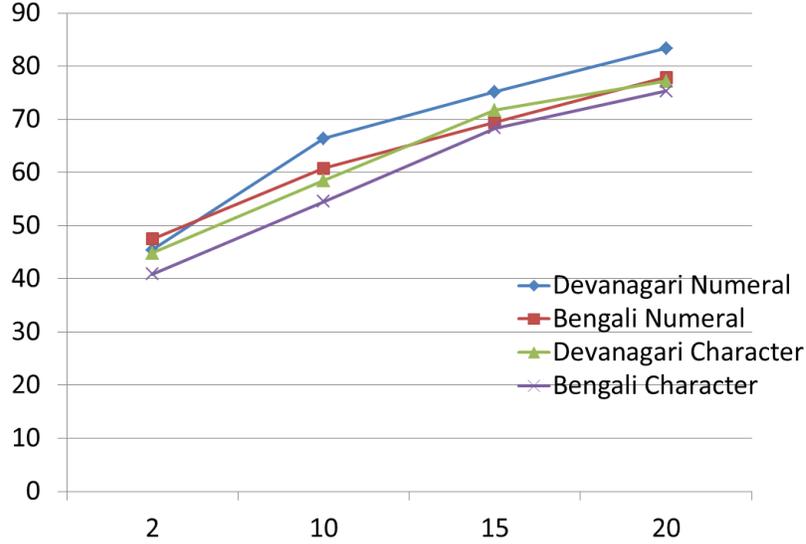
**Fig.17: Variations in accuracy with number of fonts in numeral and character recognition**

It was noted that by combining synthetic data with original handwritten data, the performance can be improved further, since the accuracy of the system depends on the variation of data used. As more handwritten data is provided with the amount of synthetic data, more variations of characters and words get generated and thus the performance of the system improves. Table III compiles the results of numeral and character recognition using different training data. When distortion of different kind like curve, sinusoidal and elliptical deformations are added to handwritten data, the accuracy of the system does not improve much.

**Table III. Recognition performance with numerals and characters using synthetic data.**

| Script | Numeral | | Character | |
|---|---|---|---|---|
| | Devanagari | Bengali | Devanagari | Bengali |
| **Synthetic Data** | 82.34% | 77.91% | 77.21% | 75.32% |
| **Handwritten Data** | 79.58% | 76.59% | 76.21% | 73.15% |
| **Synthetic + Handwritten** | 86.62% | 82.98% | 81.10% | 79.61% |

### 5.1.3. Error Analysis

On doing in-depth analysis we found out that, in Devanagari script, the system confused between the numerals 4 and 5 due to their similar forms. We also observed that the system falsely identified 9 as 1 in this script. In Bengali script, the system confused between numerals 1 and 9 and also between numerals 3 and 6. The confusion matrices on these numbers are shown in Fig.18.

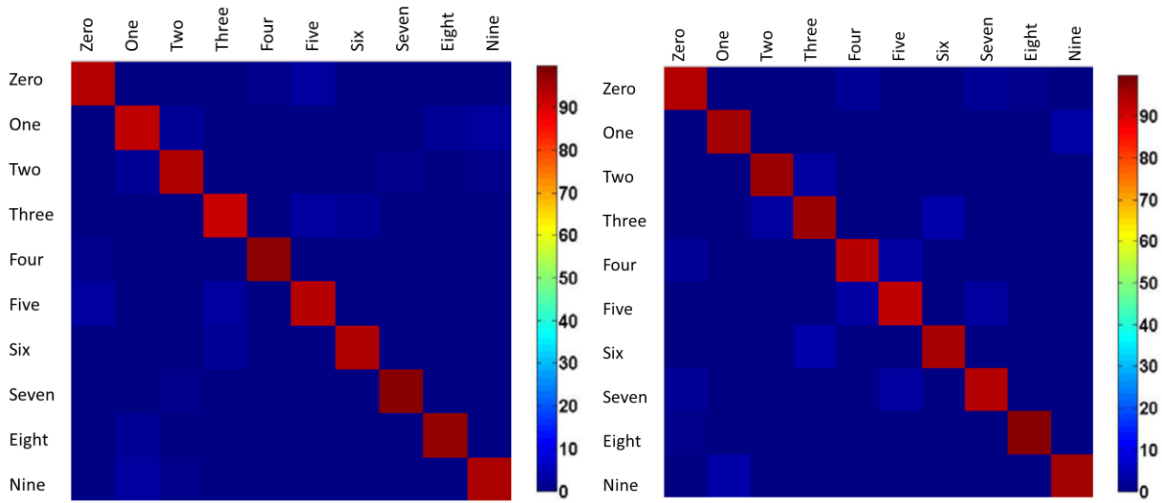

**Fig.18: Confusion matrix of numeral recognition (a)Devanagari (b)Bengali**

**5.1.4. Comparison with other System**

In [9] N. Das et al. used Principal Component Analysis (PCA) features for the recognition of numerals of Indic scripts. The training was done using handwritten numerals while we have trained our system using synthetically generated numerals as well as combining handwritten data and synthetic data. We have also extracted features using HOG [29] and PHOG [31] features. These features are fed to SVM classifier for to obtain recognition results. We observed that PHOG features provided better results than simple PCA and HOG features. The improved recognition performance in [9] was due to addition of complex Quad-tree based topological features. The comparison of results with our proposed system is given in Table IV.

**Table IV: Comparison of numeral recognition results with [12]**

| Script | PCA [12] | Present System (Synthetic only) | | | Present System (Handwritten only) | | | Present System (Synthetic + Handwritten) | | |
|---|---|---|---|---|---|---|---|---|---|---|
| | | PCA | HOG | PHOG | PCA | HOG | PHOG | PCA | HOG | PHOG |
| **Bengali** | 88.45% | 76.54% | 77.22% | 76.24% | 77.28% | 80.22% | 81.37% | 81.31% | 81.38% | 82.87% |
| **Devanagari** | 86.10% | 82.11% | 79.56% | 80.12% | 83.38% | 82.17% | 85.54% | 85.21% | 84.31% | 86.31% |

## 5.2. Word Recognition

### 5.2.1. Dataset Description

Handwritten cursive word images are considered in this experiment. For training our systems, we generated training data of Devanagari and Bengali words using our proposed approach. The words and lexicon for experiment were collected from a local news website. We applied our distortion model generation approach and produced training images of 24,000 and 25,000 from Devanagari and Bengali scripts. Fig.19 illustrates the percentage of words formed according to the length i.e. the number of characters present in the word. Table V shows the number of images used for training and testing for numerals and word recognition. A list of 1,523 Bengali words and 1,988 Devanagari words were considered in the lexicon in all experiments.

**Table V: Data details used in performing experiment for numeral and characters.**

| Script | Experiment | Training | Testing |
|---|---|---|---|
| Devanagari | Synthetic | 24000 | 6000 |
| Devanagari | Synthetic+ handwritten | 32000 | 6000 |
| Devanagari | Synthetic + handwritten (deformed) | 51000 | 6000 |
| Bengali | Synthetic | 25000 | 7263 |
| Bengali | Synthetic+ handwritten | 34000 | 7263 |
| Bengali | Synthetic + handwritten (deformed) | 54000 | 7263 |

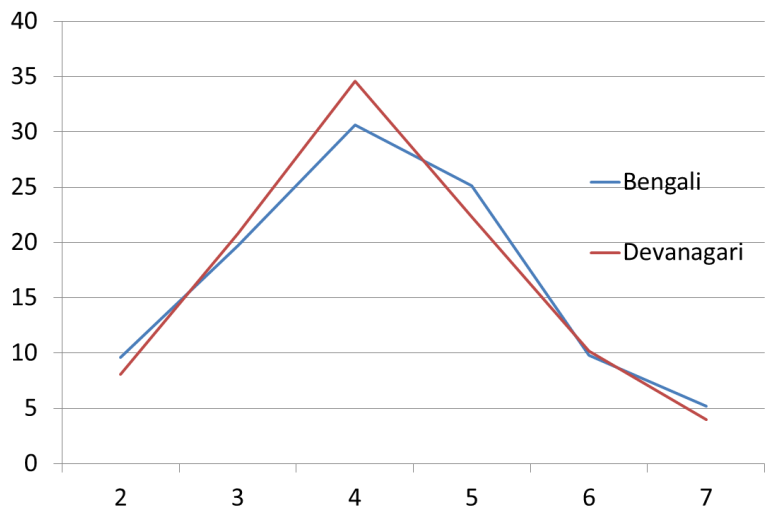

**Fig.19: Percentage of words according to their length**

### 5.2.2. Recognition Performance

After generating the training data using proposed approach (random shift distortion), handwritten words are recognized using HMM. For word recognition, first PHOG feature vectors were obtained and then these features were fed to the HMM classifier for recognition. A comparison was done with respect to the recognition obtained without and with zone segmentation. Table VI shows the recognition accuracy with synthetic, handwritten, and combining synthetic and handwritten data. Also a result was computed based on the top 5 choices for word recognition. Table VII demonstrates the top 5 recognition performance obtained in Devanagari and Bengali scripts. It is to be noted that we obtained a little higher performance using synthetic generated data than handwritten data only. It is mainly due to the large number of data and variation present in synthetic data.

**Table VI: Accuracy of full word recognition using Zone Segementation**

| Training Data Type | Scripts | Recognition Performance | |
|---|---|---|---|
| | | Without Zone Segmentation | With Zone Segmentation |
| Synthetic Data Only | Bengali | 60.47% | 74.61% |
| | Devanagari | 63.35% | 75.42% |
| Handwritten Data only | Bengali | 62.11% | 75.41% |
| | Devanagari | 66.01% | 78.31% |
| Synthetic + Handwritten Data | Bengali | 64.58% | 77.55% |
| | Devanagari | 67.75% | 79.32% |

**Table VII: Accuracy of full word recognition using synthetic words**

| Type of Training Data | Script | Recognition Accuracy | | | | |
|---|---|---|---|---|---|---|
| | | Top 1 | Top 2 | Top 3 | Top 4 | Top 5 |
| Synthetic Only | Bengali | 74.61% | 79.92% | 83.53% | 86.17% | 89.13% |
| | Devanagari | 75.42% | 80.31% | 84.13% | 87.35% | 90.17% |
| Handwritten | Bengali | 71.77% | 76.46% | 80.68% | 83.34% | 87.65% |
| | Devanagari | 72.62% | 78.53% | 83.75% | 86.53% | 89.19% |
| Synthetic + Handwritten | Bengali | 77.55% | 81.92% | 85.01% | 88.32% | 91.34% |
| | Devanagari | 79.32% | 83.67% | 86.94% | 89.97% | 92.21% |

We have compared the performance of PHOG feature with Marti-Bunke feature [25] which is used extensively for handwriting recognition. This feature consists of nine features computed from foreground pixels in each image column. Three features are used to capture the fraction of foreground pixels, the centre of gravity and the second order moment and remaining six features comprise of the position of the upper and lower profile, the number of foreground to background pixel transitions, the fraction of foreground pixels between the upper and lower profiles and the gradient of the upper and lower profile with respect to the previous column. The recognition results using Marti-Bunke feature when applied to combining synthetic and handwritten data are 74.81% and 75.23% for Bengali and Devanagari, respectively. Using PHOG, we obtained 77.55% and 79.32% results in these scripts respectively. Hence, we have used PHOG features in our experiment.

### 5.2.3. Font wise recognition results

In word recognition framework, we also perform evaluation of fonts. Some of the fonts were taken and individually fed into the recognition model and tested for accuracy obtained. The disparities in accuracy of different fonts arise due to the fact that they have different patterns and styles. By proper usage of fonts we can get a higher degree of accuracy. Fig.20 demonstrates recognition results according to font variation.

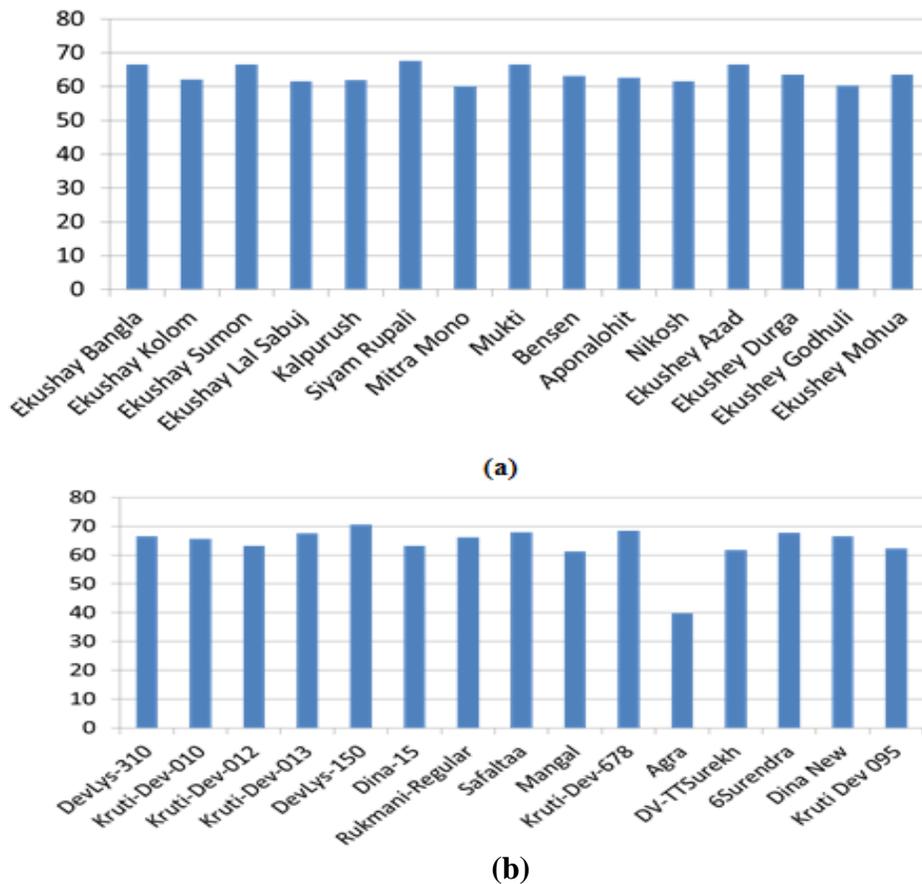

**Fig.20: Performance evaluation on handwritten test data (a) Bengali (b) Devanagari**

### 5.2.4. Recognition with different distortion model

Recognition was also performed based on the different distortion models to give an insight about the effect of the different distortion models on the recognition of the system. Fig.21 demonstrates the accuracy obtained from different distortion models like Gaussian, random shift, curved, elliptical, etc.

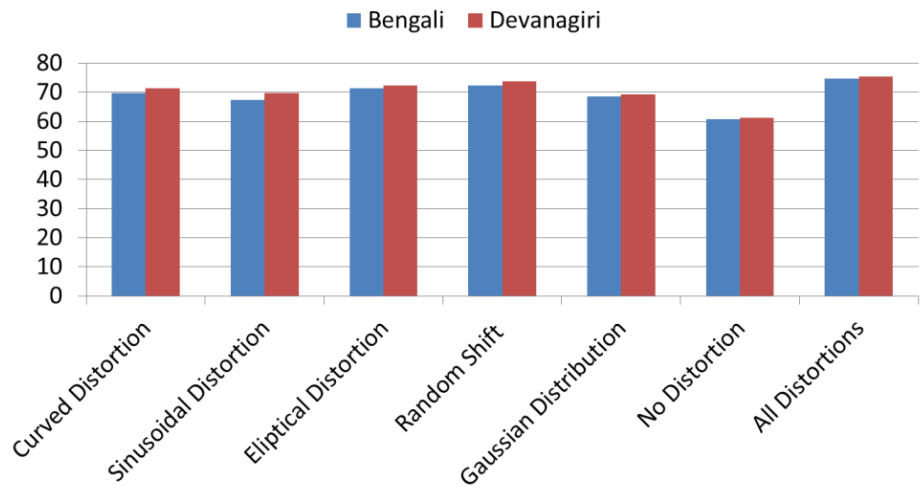

**Fig.21: Word recognition performance with different distortions.**

### 5.2.5. Parameter Evaluation

We also evaluated the parameters of HMM during experiment. Best results using PHOG features were obtained with window size of 40 times 6 and step-size of 3. An overlapping ratio of 50% was considered for feature extraction. We tested our system with various Gaussian numbers and state numbers and observed that maximum accuracy is achieved with Gaussian number 32 (See Fig.22) and state number 8 (See Fig. 23). Fig. 24 shows the scalability of system with the number of fonts used for generating training data. Fig. 25. shows the variation in accuracy for word recognition according to the number of images used for training the system. For testing the system a standard 4000 images for Devanagari and 5000 images for Bengali were used in all the cases

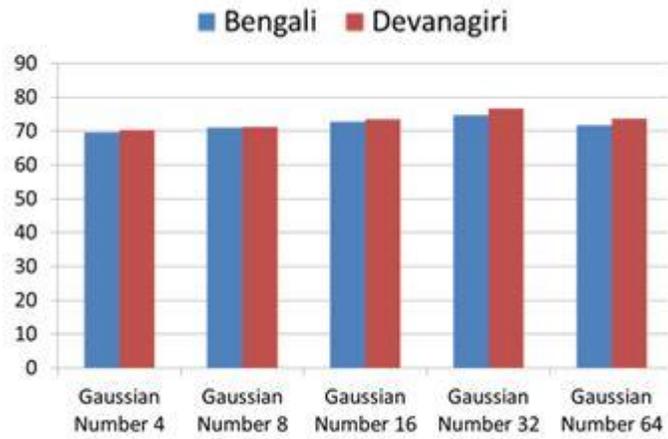

**Fig.22: Word recognition results for script**

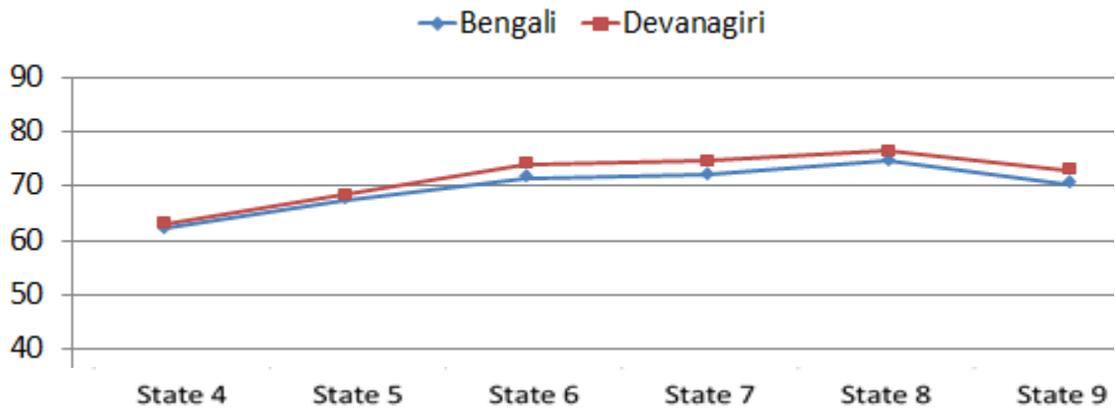

**Fig.23: Comparison of results with variation of state numbers**

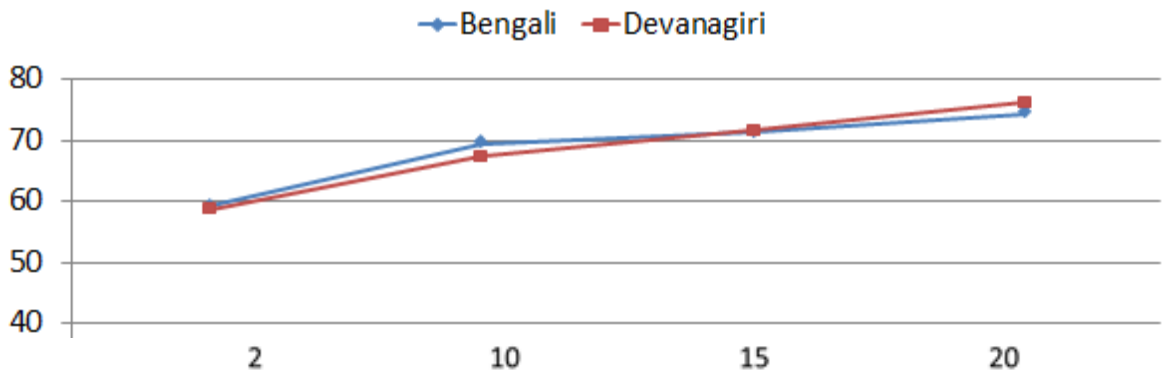

**Fig.24: shows variation accuracy with the number of fonts**

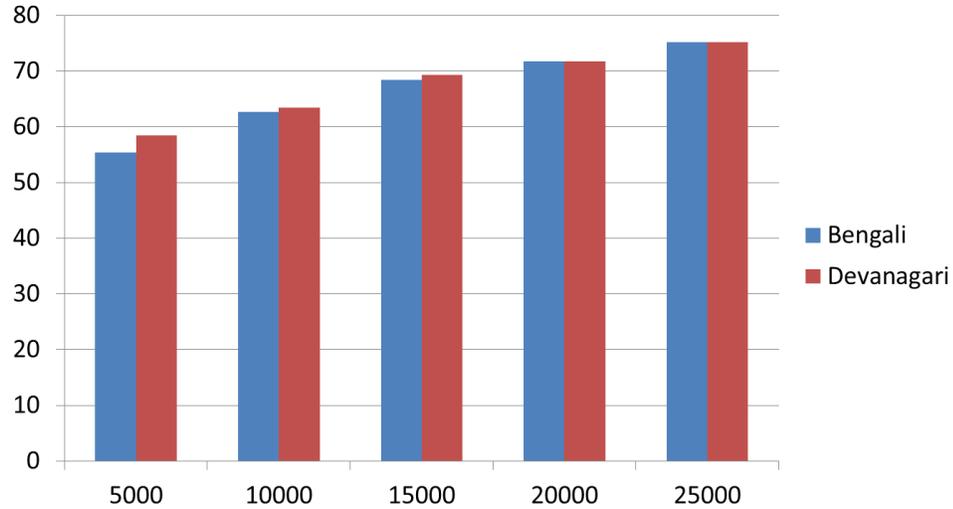

**Fig.25: Variations in accuracy with number of words used for training**

**5.2.6. Performance evaluation using Latin script**

To compare the performance, we have used IAM dataset [25] for evaluation. The testing dataset of IAM has been used to evaluate the performance. Total 34,000 words were considered for training and 6,000 for testing. Some fonts which were used for synthetic text generation were Times New Roman, Calibri, Arial, Book Antiqua, Century, etc. Cursive fonts like Monotype Corsiva, Brush Script MT, Vladimir Script, Freestyle Script, etc. were also used to generate similar to handwritten data. Fig.26 shows some synthetically generated images of Latin script. Next, PHOG feature were extracted from these text images and HMM based word recognition is performed. Table VIII shows the recognition performance using synthetic and handwritten dataset. By combining synthetic data the performance of handwritten text recognition has been improved.

In [35], authors showed an improvement of 2.69% in IAM dataset by adding synthetic data which were generated by perturbation model from handwritten data only. The authors of [23] used a subset of IAM dataset and applied synthetic data using n-tuples of characters and obtained 0.3% improvement. With our system we have obtained 3.28% improvement from using only handwritten data.

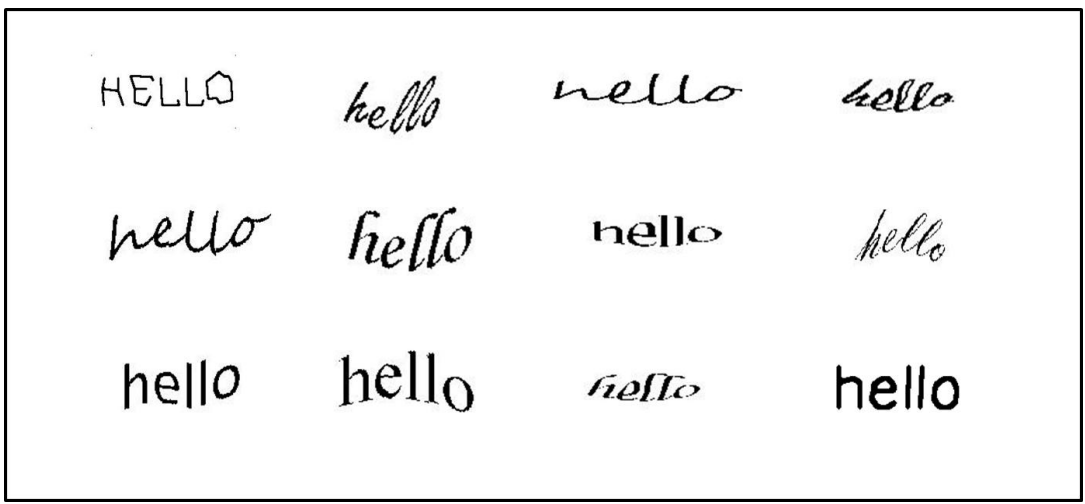

**Fig.26: Some synthetically generated images in Latin script**

**Table VIII: Accuracy of full word recognition using synthetic and handwritten words.**

| Training Data Type | Recognition Performance |
|---|---|
| Synthetic Data Only | 71.34% |
| Handwritten data Only | 73.48% |
| Synthetic + Handwritten Data | 76.76% |

## 5. Conclusion and Future Work

In this paper we introduced a novel handwriting synthesis system that greatly reduces human effort without compromising much on the efficiency of the system. Our system can be efficiently used to create both cursive and print style languages, if the fonts are styled in that manner. We considered 3 languages to present our results and demonstrated the hugely reduced human effort and improved efficiency on standard testing dataset. The proposed framework eliminates the need for the user to manually collect handwritten data to create a recognizer. This is specifically usable for scripts in which not a lot of research has been done and no formal databases have been created.

The proposed system, however, treats and weighs each font uniformly. Due to inclusion of fancier fonts not close to natural handwriting, results can be adversely affected. In [29], authors proposed a method in which computer fonts are weighed to increase recognition rates. This feature can be extended to synthetic writing generation by finding optimum weightage of fonts for training data. The system after incorporating weights can be used along with natural training data. Hence recognition performance of the hybrid system can be improved even with highly limited natural data.

During generation of dataset, we connected the junction points of the vectorized image using straight lines. We can alternatively use other functions in place of straight line, like Bezier splines or curve fitting functions, in order to produce better quality images. This can lead to a significant increase in accuracy of the system.